%% file: root.tex
\documentclass[letterpaper, 10 pt, journal, twoside]{IEEEtran}

\IEEEoverridecommandlockouts                              % This command is only needed if 
\usepackage{graphics} % for pdf, bitmapped graphics files
\usepackage{epsfig} % for postscript graphics files
\usepackage{amsmath} % assumes amsmath package installed
\usepackage{amssymb}  % assumes amsmath package installed
\usepackage{algorithm}
\usepackage[noend]{algpseudocode}
\usepackage[dvipsnames]{xcolor}

\usepackage[noadjust]{cite}
\usepackage{filecontents}
\usepackage{ulem}
\usepackage{hyperref}

\input{defs}

\author{Wenhao Yu$^{1,3}$, Jie Tan$^{1}$, Yunfei Bai$^{2}$, Erwin Coumans$^{1}$, and Sehoon Ha$^{1}$% <-this % stops a space
\thanks{Manuscript received: September, 10, 2019; Revised December, 22, 2019; Accepted January, 23, 2020.}%Use only for final RAL version
\thanks{This paper was recommended for publication by Editor Tamim Asfour upon evaluation of the Associate Editor and Reviewers' comments.} %Use only for final RAL version
\thanks{$^{1}$Jie Tan, Erwin Coumans, and Sehoon Ha are with Robotics at Google, Mountain View, CA, 94043, USA. This research was conducted during Wenhao's internship at Robotics at Google. {\tt\footnotesize \{jietan, erwincoumans, sehoonha\}@google.com}}%
\thanks{$^{2}$Yunfei Bai is with X, Mountani View, CA, 94043, USA {\tt\footnotesize yunfeibai@google.com}}%
\thanks{$^{3}$Wenhao Yu is with Georgia Institute of Technology, Atlanta, GA, 30332, USA {\tt\footnotesize wenhaoyu@gatech.edu}}%
\thanks{Digital Object Identifier (DOI): see top of this page.}
}

\title{
Learning Fast Adaptation with Meta Strategy Optimization
}

\begin{document}

\maketitle
% \thispagestyle{empty}
% \pagestyle{empty}

%%%%%%%%%%%%%%%%%%%%%%%%%%%%%%%%%%%%%%%%%%%%%%%%%%%%%%%%%%%%%%%%%%%%%%%%%%%%%%%%
\begin{abstract}
The ability to walk in new scenarios is a key milestone on the path toward real-world applications of legged robots. In this work, we introduce Meta Strategy Optimization, a meta-learning algorithm for training policies with latent variable inputs that can quickly adapt to new scenarios with a handful of trials in the target environment. The key idea behind MSO is to expose the same adaptation process, Strategy Optimization (SO), to both the training and testing phases. This allows MSO to effectively learn locomotion skills as well as a latent space that is suitable for fast adaptation. We evaluate our method on a real quadruped robot and demonstrate successful adaptation in various scenarios, including sim-to-real transfer, walking with a weakened motor, or climbing up a slope. Furthermore, we quantitatively analyze the generalization capability of the trained policy in simulated environments. Both real and simulated experiments show that our method outperforms previous methods in adaptation to novel tasks.
\end{abstract}


\input{introduction.tex}

\input{related_works.tex}

\input{method.tex}

\input{experiments.tex}
\input{conclusion.tex}

\addtolength{\textheight}{-1cm}   % This command serves to balance the column lengths

%%%%%%%%%%%%%%%%%%%%%%%%%%%%%%%%%%%%%%%%%%%%%%%%%%%%%%%%%%%%%%%%%%%%%%%%%%%%%%%%

%%%%%%%%%%%%%%%%%%%%%%%%%%%%%%%%%%%%%%%%%%%%%%%%%%%%%%%%%%%%%%%%%%%%%%%%%%%%%%%%


\section*{ACKNOWLEDGMENT}

The authors gratefully thank Tingnan Zhang, Karol Hausman, Benjamin Eysenbach, the locomotion team at Google Robotics, and the anonymous reviewers for valuable discussion and suggestions.

%%%%%%%%%%%%%%%%%%%%%%%%%%%%%%%%%%%%%%%%%%%%%%%%%%%%%%%%%%%%%%%%%%%%%%%%%%%%%%%%

\bibliographystyle{IEEEtran}
\bibliography{IEEEabrv,reference}

\end{document}

%% file: defs.tex
\newcommand{\cmt}[1]{}

\newcommand{\newtext}[1]{{#1}}
\newcommand{\original}[1]{}

\newcommand{\figref}[1]{Figure~(\ref{fig:#1})}

\long\def\ignorethis#1{}

\newcommand{\etal}{{\em{et~al.}\ }}

\newcommand{\vc}[1]{\ensuremath{\mathbf{#1}}}

\newcommand{\pctab}{\hspace{0.2in}}

%% file: introduction.tex
\section{Introduction}

% Agile walking in novel environments comes naturally to human beings and animals. However, current legged robots still lack this key skill that prevents them from being applied to real-world applications. Teaching a robot how to adapt to new environments requires exposing it to a large variety of environments and a large amount of experiences, which makes computer simulation an attractive tool for learning locomotion policies for legged robots. In deed, recent developments in deep reinforcement learning and transfer learning have achieved successful locomotion policies on real legged robot, where the training happens mostly in computer simulation. However, the resulting policy is usually limited to a single target environment, i.e. the real robot, and may not work if the target environment changes notably, for example if the robot is placed on a different type of terrain.

% To equip the learned policy with the ability to handle new environments, a common choice is to train a single robust policy that can handle a large range of environments in the simulation and hope that it can be robust to novel situations. Though with careful system identification, a robust policy could be successfully applied on a real hardware, the robustness comes at the cost of flexibility: it is usually difficult to quickly adjust the behavior of these policies in order to adapt to environments that requires a distinct control strategy from the learned one.

\IEEEPARstart{H}{uman} beings and animals have a natural ability to adapt their motor skills to novel situations. Robots in the real world also often encounter unexpected tasks and environments, such as manipulating unseen objects or walking on unstructured terrains. Many of state-of-the-art robots still lack this capability to adapt, which prevents them from real-world deployment. 

Recent advances in deep reinforcement learning (deep RL) shed light on developing effective motor skills in challenging situations \cite{schulman2017proximal, lillicrap2015continuous, mnih2016asynchronous}. However, the policy found by deep RL is usually limited to a single scenario and may not work if the target environment changes notably. One of the common techniques to overcome this generalization issue is to train a single policy that can handle a wide range of situations by exposing it to many random scenarios, so-called \emph{domain randomization} (DR) \cite{tobin2017domain}. DR is known to be effective for generating a single robust policy across different scenarios, which is suitable for some problems such as sim-to-real transfer. However, DR trades optimality for robustness: the policies learned by DR is not optimal under any situation. Another popular approach is \emph{meta reinforcement learning} (meta-RL) \cite{rakelly2019efficient, finn2017model} that aims to solve a new task within a few iterations by training adaptation over a distribution of tasks. However, existing meta-RL methods are mostly effective for adapting to different reward functions, while are in general less effective for adapting to challenging control problems where the dynamics are changed \cite{yang2019norml}.
%However, it is not straightforward \yunfeibai{maybe more explicit about the challenge here?} to develop a robust meta-RL algorithm that can handle challenging motor control problems with fast adaptation to unseen environments.
\begin{figure}
    \centering
    \includegraphics[width=0.48\textwidth]{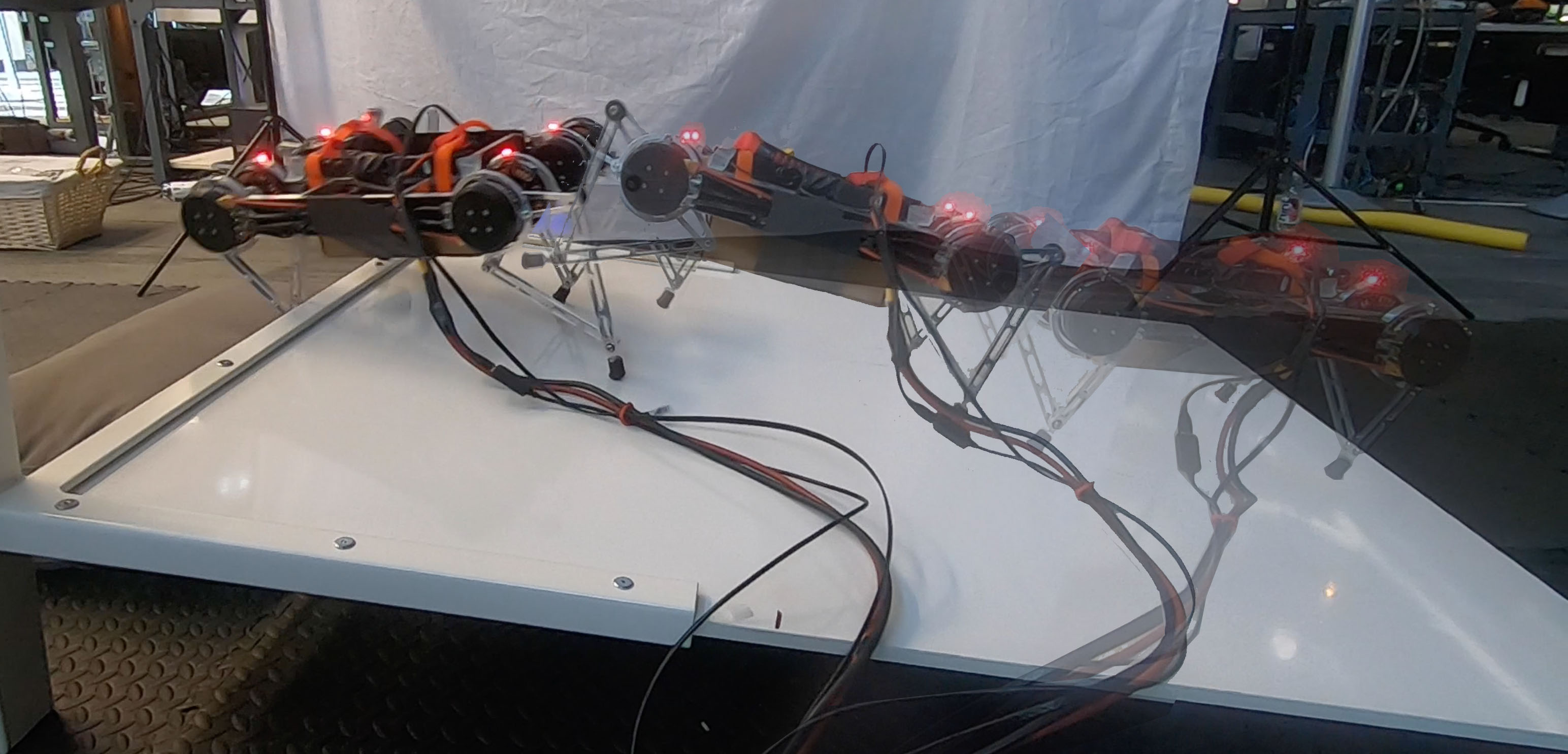}
    \caption{\small Policies trained using our method adapts to sloped surface on the real quadruped robot in $15$ episodes. During training in simulation, it has only seen flat ground.}
    \vspace{-2mm}
	\label{fig:teaser}
\end{figure}
In this work, we aim to develop a meta-RL algorithm that can quickly adapt the behavior of the trained policies to novel reward functions and dynamics that are not examined during training. We extend the idea of Strategy Optimization (SO) \cite{yu2018policy} that trains a policy modulated by a latent variable to exhibit versatile behaviors. During the evaluation, the latent variable is adapted directly on the hardware using a sampling-based optimization method. The key idea behind our proposed method, Meta Strategy Optimization (MSO), is to expose the learning agent to the same Strategy Optimization process during both training and testing phases. This meta-training allows the agents to learn a better latent policy space that is suitable for fast adaptation to new situations.

We demonstrate our proposed algorithm on training locomotion policies for the Ghost Robotics Minitaur \cite{kenneally2016design}, a quadruped robot. Our algorithm can successfully train locomotion policies that can be applied to the real hardware by adjusting its simulation-acquired behavior (Figure \ref{fig:teaser}). In addition, we design two adaptation tasks for the real robot, walking with a weakened leg and climbing a slope, and a set of additional tasks in a simulated environment. We show that MSO is extremely data efficient ($\leq 15$ rollouts or $75$ seconds of data) to adapt the policies to novel situations in the target environment. We compare our method to two baseline methods: domain randomization \cite{TanRSS18} and strategy optimization with a projected universal policy \cite{yu2019sim}. Our results show that MSO outperforms both baselines in the simulated and the real environments.

%% file: related_works.tex
\section{Related Works}

%In this work, we study the problem of legged robots learning to walk in novel environments. In order to minimize the interactions required on the hardware, we resort to computer simulation software for providing synthetic environments for the robot to learn the locomotion skills and transfers the skill to the real hardware. There are two key challenges in doing this: 1) make sure the locomotion policy trained in simulation can succeed on the real hardware, and 2) quickly adapt a trained policy to novel environments. In this section, we survey prior works that are relevant to each of the two problems and discuss the differences to our proposed algorithm.

\subsection{Sim-to-real transfer for legged locomotion}

Recent developments in deep reinforcement learning (Deep RL) have enabled training locomotion policies for legged robots with high dimensional observation and action spaces and challenging dynamics \cite{schulman2017proximal, lillicrap2015continuous, mnih2016asynchronous}, which demonstrate an attractive path toward automatically acquiring motor skills for robots. However, the sample complexity and potential safety concerns prevents deep RL from being applied directly on the hardware, while the discrepancies between computer simulation and real world, also known as the Reality Gap \cite{neunert2017off}, makes a simulation trained policy unlikely to work on the real robot.

Researchers have proposed a variety of techniques to enable a policy trained in simulation to be transferred to the real robot \cite{TanRSS18, yu2019sim, hwangbo2019learning, peng2018sim, learndexmanipulation18, hanna2017grounded}. One important strategy is to improve the computer simulation to better match the real robot dynamics \cite{TanRSS18, hwangbo2019learning, hanna2017grounded}. For example, Tan et al. \cite{TanRSS18} improved the actuator dynamics by identifying a nonlinear torque-current relation and demonstrated successful transfer of locomotion policies for a quadruped robot. In this work, we leverage the model parameters and the nonlinear actuator model identified by Tan et al. \cite{TanRSS18} for the quadruped robot. However, improving the simulation model alone does not allow the policy to be transferred to notably different dynamics or tasks.

Another important technique for sim-to-real transfer is to train control policies that are robust to a range of simulated environments and sensor noises. Different techniques have been proposed to train robust policies, such as domain randomization \cite{tobin2017domain,peng2018sim,learndexmanipulation18, yan2019data}, adversarial perturbation \cite{pinto2017robust}, and ensemble models \cite{Mordatch,Lowrey}. Though training a policy with pure domain randomization may transfer to the real robot, it usually assumes that the training dynamics are not too far from the target dynamics. As shown in our experiments, domain randomization alone fails to transfer if the reality gap is large. In addition, without a mechanism to adjust the policy behavior, these policies cannot quickly adapt to cases where the reward function is changed.

\newtext{In vision community, researchers have also investigated the problem of sim-to-real transfer to overcoming the discrepancies between rendered and real images. One of the most successful methods is domain adaptation \cite{james2019sim, bousmalis2018using, fang2018multi}. It trains a generative model to transform the observations from the source domain to the target domain. In this work, the main challenge is to adapt policies for dynamic changes, which is very different from visual changes.}

\subsection{Adapting control policy to novel tasks}

To adapt to new reward functions or dynamics, it is necessary that the controller can modify its behavior according to the real-world experience. Existing works in this line of research can be roughly divided into two categories: model-free adaptation method and model-based adaptation method.

In model-free adaptation method, the control policy is directly adjusted according to experience from the target environment. One class of such method is the gradient-based meta learning approach \cite{finn2017model, houthooft2018evolved, rothfuss2018promp, yang2019norml}, where the goal is to train policies that can be quickly adapted by gradient-based optimization methods during test time. Gradient-based meta learning methods have been demonstrated on adapting to novel reward function and are universal in theory \cite{finn2017meta}. However, it is in general less effective for adapting to novel dynamics. No-Reward Meta Learning (NoRML) \cite{yang2019norml} addressed this issue by meta-learning an advantage function and an offset in addition to the policy parameters. NoRML has demonstrated effective adaptation to unseen dynamics in simulation. However, it has yet been demonstrated on real robots.

In contrast to gradient-based method, latent space based adaptation method encodes the training experience into a latent representation \cite{rakelly2019efficient, YuRSS17, yu2018policy, duan2016rl, james2018task}. The policy is then fine-tuned when a new environment is presented. Most methods in this class try to infer the latent input using observations from the target environment. For example, Yu et al. \cite{YuRSS17} conditioned the policy on the physics parameters of the robot, and trained a separate prediction model that estimates the physics parameters given the history of observations and actions. These methods can potentially adapt to changes in environments in an online fashion. However, when the dynamics changes significantly, the inference model may produce non-optimal latent inputs. As a result, most works have been demonstrated in simulated environments only. 

Instead of training an inference model, researchers have also proposed methods that directly optimizes the latent input to the policy in the target environment \cite{yu2019sim, yu2018policy}. As the latent space that the policy is conditioned on is usually low dimensional, it is possible to use sampling-based optimization methods such as CMA-ES \cite{hansen1995adaptation}, or Bayesian Optimization \cite{mockus2012bayesian} to find the best latent input that achieves the highest performance. Such methods have been successfully applied to learning locomotion policies for a biped robot \cite{yu2019sim}\original{ and adapting to novel environments for a hexapod robot \cite{cully2015robots}}. Our method extends this line of research by matching the process of optimizing latent input during training and testing. We demonstrate that by doing this, we learn a better latent space that is suitable for fast adaptation.

\newtext{Another related line of work is to define a space of robot behaviors, and then optimize on the real robot \cite{cully2015robots, rai2018bayesian}. For example, Cully et al. demonstrated fast adaptation on a hexapod robot, by precomputeing a behavior-performance map and using Bayesian Optimization to search the map for the optimal controller when the robot is damaged \cite{cully2015robots}. Rai et al. also used Bayesian Optimization to optimize locomotion controllers for the ATRIAS Biped with less than 10 trials on the robot \cite{rai2018bayesian}. These approaches usually require designing a low-dimensional behavior space using domain knowledge. In contrast, our method applies meta learning, which leverages many different dynamic environments in training, to implicitly shape the lower-dimensional search space for the on-robot optimization.}

Model-based adaptation method, on the other hand, adapts the dynamics model learned in source domain and extracts the control policy using methods such as model-predictive control (MPC) \cite{nagabandi2018learning, tanaskovic2013adaptive, aswani2012extensions, manganiello2014optimization, lenz2015deepmpc}. These methods have the advantage of being data efficient \newtext{and can naturally adapt to changes in the environment online}. However, \newtext{performing inference of the optimal action in an MPC style is more computationally expensive, and }the learned dynamics model usually uses the full state of the robot, which requires additional instruments such as a motion capture system.

%% file: method.tex
%%%%%%%%%%%%%%%%%%%%%%%%%%%%%%%%%%%%%%%%%%%%%%%%%%%%%%%%%%%%%%%%%%%%%%%%%%%%%%%%
\section{Background}

We represent the problem of legged locomotion as a Markov Decision Process (MDP): $(S, A, p, r, \rho_0)$, where $S$ is the state space of the robot, $A$ is the action space, $p: S \times A \mapsto S$ is the transition function, $r: S \mapsto R$ is the reward function and $\rho_0$ is the initial state distribution.
The goal of reinforcement learning is to find a policy $\pi: S \mapsto A$, such that it maximizes the expected accumulated reward over time under the transition function $p$:

\begin{equation}
    J_p(\pi) = \mathbb{E}_{\mathbf{s}_0, \mathbf{a}_0, \dots, \mathbf{s}_T} \sum_{t=0}^{T} \gamma^t r(\mathbf{s}_t, \mathbf{a}_t),
\end{equation}
 where $\mathbf{s}_0 \sim \rho_0$, $\mathbf{a}_t \sim \pi(\mathbf{s}_t)$ and $\mathbf{s}_{t+1}\sim p(\mathbf{s}_t, \mathbf{a}_t)$. In deep reinforcement learning, the policy is usually parameterized by a neural network with weights $\theta$ and the policy is denoted as $\pi_\theta$.

\begin{figure}
    \centering
    \includegraphics[width=0.48\textwidth]{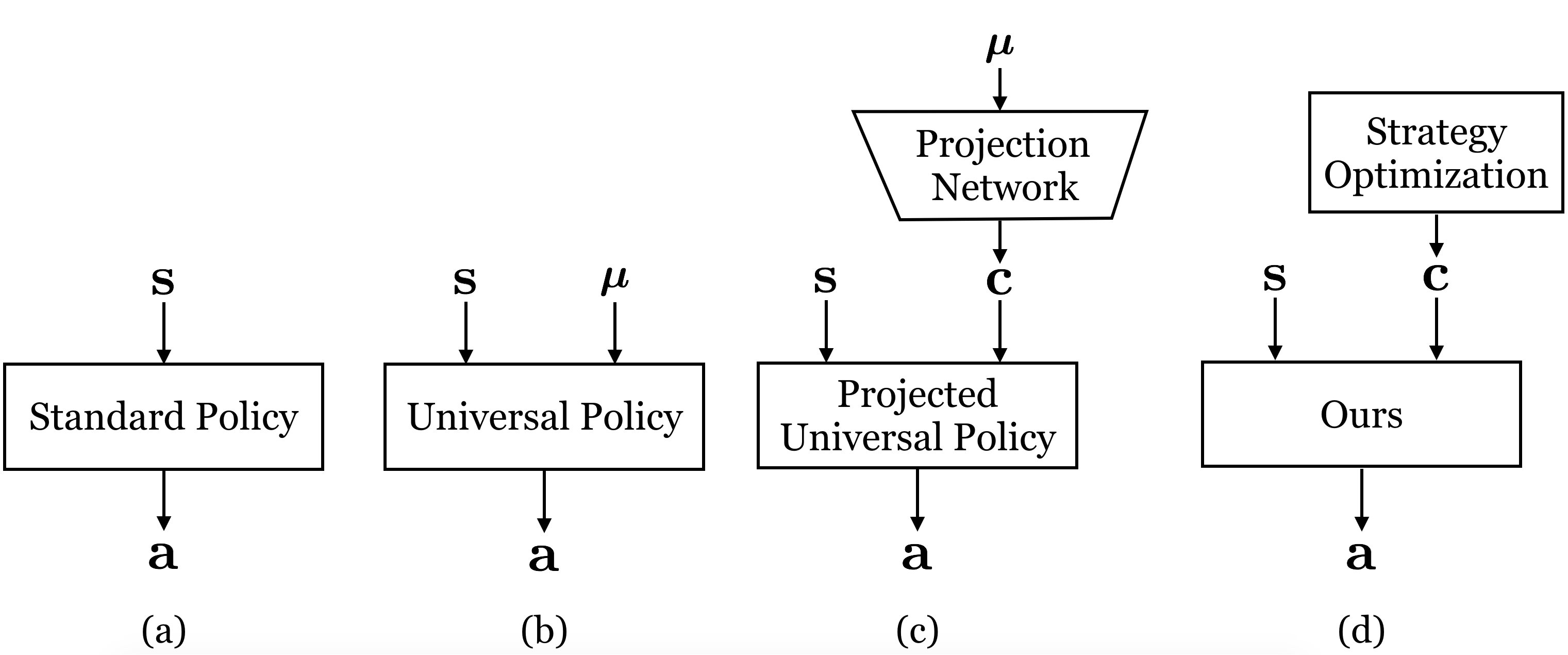}
    \vspace{-5mm}
    \caption{\small (a) The standard feed-forward policy (b) a universal policy (UP) that takes high-dimensional physics parameters as additional inputs (c) a projected universal policy (PUP) that takes the compressed physics parameters, contexts. (d) our method is designed to quickly adapt low-dimensional context variables to new situations.}
	\label{fig:policies}
	\vspace{-2mm}
\end{figure}

Strategy Optimization (SO) \cite{yu2018policy} extends the standard policy learning by training a universal policy (UP) that is conditioned on physics parameters $\boldsymbol{\mu}$ of the simulated robot: $\pi_\theta(\mathbf{s}, \boldsymbol{\mu})$ (\figref{policies}b). Under the assumption that we have access to the true physics parameters (e.g. in simulated environments), we can train a universal policy with any standard reinforcement learning algorithm by treating $\boldsymbol{\mu}$ as part of the observations. The trained universal policy will change its behaviors with respect to different physics parameters $\boldsymbol{\mu}$, thus a policy with a particular physics parameter input $\pi_\theta(\mathbf{s}, \boldsymbol{\mu})$ can be treated as \textit{strategy}.

In order to transfer the trained policy to the real world, SO solves the following optimization directly on the hardware:
\begin{equation}
\label{eq:so_obj}
    \boldsymbol{\mu}^* = \arg\max_{\boldsymbol{\mu}} J_{real}(\boldsymbol{\mu}, \theta),
\end{equation}
where $J_{real}(\boldsymbol{\mu}, \theta)$ denotes the performance of the strategy $\pi_\theta(\mathbf{s}, \boldsymbol{\mu})$ on the real robot. As the search space is significantly smaller than the network weight space, it permits the use of sampling-based optimization methods such as CMA-ES \cite{hansen1995adaptation} or Bayesian Optimization \cite{mockus2012bayesian}, which can better handle noisy objectives such as the one used in RL than gradient based methods \cite{varelas2019benchmarking}. To further reduce the search space during the transfer, Yu \etal proposed a projected universal policy (PUP, \figref{policies}c)~\cite{yu2019sim}, which projects the physics parameters $\boldsymbol{\mu}$ to a lower dimensional latent space of context variables $\mathbf{c}$ (usually $2$-$3$ dimensional). At the learning phase, PUP takes the robot observation and physics parameters as input, while during strategy optimization PUP directly optimizes the low-dimensional context variables instead of the physics parameters.

Strategy optimization with projected universal policy (SO-PUP) has demonstrated successful sim-to-real transfer for biped locomotion problems. However, during the training of SO-PUP, the latent space of context variables are acquired through the projection network that has never experienced the adaptation process before. As we demonstrate in our experiments \ref{sec:exp}, this mismatch between training and testing phases leads to a learned latent space that is not in favor of fast adaptation.

\section{Meta Strategy Optimization}

\begin{algorithm}[b]
 \caption{Meta Strategy Optimization}
   \label{alg:MSO}
\begin{algorithmic}[1]
\State Randomly initialize policy weights $\theta_1$. \;
\For{$t=1:k$}
\State Sample $n$ tasks $\{\boldsymbol{\mu}_i|i=1,\dots, n\}$. \;
\State For each $\boldsymbol{\mu}_i$, solve Eq. \ref{eq:approx_meta_opt1} with $\theta_t$ and obtain $\boldsymbol{c}_{\boldsymbol{\mu}_i,t}$.\;
\For{$j=1:h$}
\State Randomly sample a pair of ($\boldsymbol{c}_{\boldsymbol{\mu},t}$, $\boldsymbol{\mu}$). \;
\State Collect rollouts with $p_{\boldsymbol{\mu}}$ and $\pi_{\theta_t}(\mathbf{s}, \mathbf{c}_{\boldsymbol{\mu},t})$.
\State Obtain $\theta_{t+1}$ by solving Equation \ref{eq:approx_meta_opt2}.
\EndFor
\EndFor
\Return{$\pi_{\theta_k}$}
\end{algorithmic}
\end{algorithm}

In this work, we present Meta Strategy Optimization (MSO), a meta-learning algorithm that learns a latent variable conditioned policy on a large variety of simulated environments and can quickly adapt the trained policy to novel reward and dynamics with a few episodes of data from the target environment. The key idea behind MSO is that we adopt the same adaptation process to obtain the latent input to the policy during both training and testing. Therefore, our policy directly takes context variables $\mathbf{c}$ as inputs~(\figref{policies} (d)).

We solve the following optimization problem during training in simulation:
\begin{equation}
\label{eq:meta_opt}
    \theta^* = \arg\max_{\theta} \mathbb{E}_{\boldsymbol{\mu}}[ \max_{\boldsymbol{c}}J_{\boldsymbol{\mu}}(\boldsymbol{c}, \theta)],
\end{equation}
where $\theta$ is the weight of the policy network, $J_{\boldsymbol{\mu}}(\boldsymbol{c})$ is the performance of the strategy $\pi_\theta(\mathbf{s}, \boldsymbol{c})$ when the physics parameters are $\boldsymbol{\mu}$. Note that we refer $\boldsymbol{\mu}$ to the physics parameters for clarity and consistency to previous works. However, one can easily extend it to include parameters from other components of the MDP such as the reward function.

Directly solving Equation \ref{eq:meta_opt} is challenging for two reasons. First, the objective term involves strategy optimization inside the expectation, which makes it difficult to compute the gradient with respect to the policy parameters $\theta$. Second, every single evaluation of the policy parameters $\theta$ involves performing SO to get the optimal strategy (Equation \ref{eq:approx_meta_opt1}), which increases the computational cost significantly. %\yunfeibai{c was used as context viable above but is described as optimal strategy here} \yunfeibai{it is unclear what ``inner loop" is referred to here.} \wenhao{modified}

\newtext{We propose a practical algorithm by observing that the optimization problem in Equation \ref{eq:meta_opt} can be written as: 
\begin{equation}
\label{eq:meta_opt_expand}
    \theta^*, \boldsymbol{c}(\boldsymbol{\mu})^* = \arg\max_{\theta, \boldsymbol{c}(\boldsymbol{\mu})} \mathbb{E}_{\boldsymbol{\mu}}[ J_{\boldsymbol{\mu}}(\boldsymbol{c}(\boldsymbol{\mu}), \theta)],
\end{equation}
where $\boldsymbol{c}(\boldsymbol{\mu})$ is a mapping from the tasks $\boldsymbol{\mu}$ to its corresponding latent variable. We can then solve the optimization problem in an approach similar to Coordinate Descent \cite{wright2015coordinate}, where we alternate between optimizing the latent variables $\boldsymbol{c}(\boldsymbol{\mu})$ and the policy network parameters $\theta$:
} 

\original{We propose a practical algorithm for solving Equation \ref{eq:meta_opt} by making the following assumption: the changes in the optimal latent input $\mathbf{c}^*$  from SO are small if the changes in the policy network weights $\theta$ are also small. As a result, we can approximately solve Equation \ref{eq:meta_opt} by interleaving the optimization of $\theta$ and $\mathbf{c}$:}

\begin{align}
\label{eq:approx_meta_opt1}
    \boldsymbol{c}_{\boldsymbol{\mu}, t} =& \arg\max_{\boldsymbol{c}} J_{\boldsymbol{\mu}}(\boldsymbol{c}, \theta_t)\\
    \label{eq:approx_meta_opt2}
    \theta_{t+1} =& \arg\max_{\theta} \mathbb{E}_{\boldsymbol{\mu}}[ J_{\boldsymbol{\mu}}(\boldsymbol{c}_{\boldsymbol{\mu},t}, \theta)],
\end{align}
where $t$ is the iteration number.

Algorithm \ref{alg:MSO} describes the MSO algorithm in more details. For each iteration of policy learning, we first sample a set of $n$ tasks from the simulator and perform strategy optimization to obtain the current best strategies for these tasks. We then perform $h$ steps of policy updates with the fixed set of task-strategy pairs. In our experiments, we use $n=5$ and $h=30$.  

By computing the latent variable $\mathbf{c}$ using strategy optimization, MSO avoids the need to compute a projection from $\boldsymbol{\mu}$ to $\mathbf{c}$ and thus can handle tasks with larger dimensions than SO-PUP. More importantly, by matching the process of obtaining the latent variable during training and testing, MSO can \original{potentially learn}\newtext{implicitly shape} a latent space \newtext{of control behaviors} that is more suitable for strategy optimization when adapting to novel scenarios.

%% file: experiments.tex
\section{Experiments}
\label{sec:exp}
We aim to answer the following questions in our experiments: 1) Does MSO achieve better performance than the baseline methods DR \cite{TanRSS18} and SO-PUP \cite{yu2019sim} in adapting to new dynamics and rewards? 2) Does MSO train policies that can be successfully transferred to real robots and adapt to novel scenarios in the real world? 3) Is MSO sensitive to the specific choice of hyper-parameters? \newtext{4) Does MSO achieve better performance on adapting to new dynamics than gradient-based meta learning algorithms?} To answer these questions, we design a set of experiments in both simulation and real-world. Videos of our results can be seen in the supplement video \footnote{Video available at:  \url{https://www.youtube.com/watch?v=Mm3IIEZ0-Nw}}.

\subsection{Experiment setup}

We use Minitaur from Ghost Robotics \cite{kenneally2016design} as the robot platform to evaluate our algorithm. Minitaur has eight direct-drive actuators, two on each leg. In this work, we use a Proportional-Derivative controller (P gain is $0.5$ and D gain is $0.005$) to track the desired motor positions, which is the output of the policy. Minitaur is equipped with motor encoders to read the motor angles and an IMU sensor to estimate the orientation and angular velocity of the robot body. The robot is controlled at a frequency of $50$ Hz.

We build a physics simulation of the Minitaur in PyBullet~\cite{pybullet}, a Python module that extends the Bullet Physics Engine. Our simulator incorporates the actuator model \cite{TanRSS18}, but we do not perform a thorough system identification for its parameters. As shown in our experiments, a na\"ive domain randomization technique does not give us a transferable policy directly.

The observation space of the robot consists of the current motor angles, the roll, pitch of the base, as well as their time derivatives. We design a reward function that encourages the robot to move forward:

\begin{equation}
    r = clip((\mathbf{p}_n - \mathbf{p}_{n-1}) \cdot \mathbf{d} / dt, -\bar{v}, \bar{v}),
\end{equation}
where $\mathbf{p}_n$ denotes the position of the robot base at timestep $n$, $\mathbf{d}$ is the desired moving direction, $dt$ is the control timestep, and $\bar{v}$ is a velocity threshold for safety reasons. We use $dt=0.02$s and $\bar{v} = 1$m/s in our experiments. Each episode of simulation has a maximum horizon of $250$ steps ($5$s). The episode is terminated early if the robot falls, determined by the roll and pitch angles of the base.

We represent the locomotion policy using a feed-forward neural network with two hidden layers, each consists of $64$ neurons. We use Augmented Random Search (ARS), a policy optimization algorithm, for training the locomotion policy in simulation \cite{mania2018simple}. At each iteration, ARS samples $d$ random perturbations of $\theta$ and estimates the policy gradient along the best performing perturbation directions using finite differences. We refer the readers to the original paper for more details. In our experiments, we sample $92$ perturbations for each iteration and use the top $23$ perturbations to update the policy weights. Although ARS has only been demonstrated for training linear policies, we find it also effectively in training neural network policies. We choose ARS because it can better leverage large scale computational resource, though MSO can also be applied to other on-policy RL algorithms such as PPO \cite{schulman2017proximal}. We use Bayesian Optimization to perform SO and limit the maximum episode number to $25$ during training.% \sehoon{Do we have $25$ for training and $15$ for testing? If yes, we need justification.} \wenhao{Yes, they are mismatched.. I put some justifications in the paragraph below.}

We compare MSO to two baselines: domain randomization (DR) \cite{TanRSS18} and strategy optimization with projected universal policy (SO-PUP) \cite{yu2019sim}. We run ARS for $1500$ iterations for all methods and we use a two-dimensional latent space for MSO and SO-PUP. Table \ref{tbl:randomization_range} shows the physics parameters and their corresponding range we use during training. During our experiments on the hardware, we find that $15$ episodes are sufficient to achieve successful adaptation. Thus we choose $15$ episodes during testing for both MSO and SO-PUP. To reduce the influence of the stochastic learning process, we train five policies for each method. Each trained policy is then evaluated on $1,500$ sampled tasks from the designed task distributions for all simulated adaptation experiments (Section \ref{ssec:tasks}).

\newtext{We further compare MSO to two gradient-based meta learning algorithms: Model-Agnostic Meta Learning (MAML) \cite{finn2017model} and No-Reward Meta Learning (NoRML) \cite{yang2019norml} on a simulated Hopper robot\footnote{\newtext{We use the implementation of MAML and NoRML from https://github.com/google-research/google-research/tree/master/norml. The Hopper environment was modeled and simulated using Dart \cite{dart2018}, and can be found here: https://github.com/DartEnv/dart-env.}}. During training of all methods, we vary the ground friction in $[0.1, 1.0]$, and the weight of the torso in $[2, 15]kg$. During testing, we evaluate the performance of the policy with an extended range ($[0.1, 1.9]$ for ground friction and $[2, 28]kg$ for torso weight) to test the generalization performance. For MSO, we run ARS for $600$ iterations, each with $32$ perturbations. We use the top $8$ perturbations for updating the policy. The rest of the hyper-parameters are the same as the ones in Minitaur experiments. For MAML and NoRML, we run the policy update for $1000$ iterations and use the default hyper-parameters for the algorithms. We allow $25$ episodes during adaptation for all three methods. The results can be found in Section \ref{maml_compare}. }

\subsection{Adaptation tasks}

\label{ssec:tasks}

We design the following tasks on the real robot to evaluate the performance of MSO:

\textbf{1) Sim-to-real transfer.} The first task is to transfer the policy trained in simulation to the real Minitaur robot. Although we use the nonlinear actuator model from Tan et al. \cite{TanRSS18}, the reality gap in our case is still large as we use a different version of Minitaur and we do not perform additional system identification.

\textbf{2) Weakened motors.} It is common for real robots to experience motor weakening, e.g. due to over heating. In this task, we test the ability of MSO to adapt to weakened motors by setting the P gain to $0.2$ for the two motors on the front right leg of Minitaur. Such strength reduction ($60\%$) is beyond the range that the policy has seen during training. 

\textbf{3) Climbing up a slope.} In this task, we place the robot on a slope of about $10$ degrees constructed by a white board and task the robot to climb up the hill. This is a challenging task because during training in the simulation the robot has only seen flat ground. 

In addition, we design the following tasks in simulation for a more comprehensive analysis of the adaptation performance of MSO:

\textbf{1) Extended randomization.} In this task, we sample dynamics from the same set of parameters used in training (Table \ref{tbl:randomization_range}), but with an extended range that is $\sim 30 \%$ wider. We also reject samples that lie within the training range to focus on generalization capability. This gives us a large space of testing dynamics that have not been seen during training.

\textbf{2) Climbing up slopes.} We also evaluate MSO for climbing up a hill in simulated environments. We randomize the angle of the slope in $[5, 20]$ degrees during evaluation. 

\textbf{3) Motor offset.} One of the common defects of actuators is that the zero position is wrong. We evaluate the ability of MSO to adapt to such issues in this task. Specifically, we add an offset sampled in $[-35, 35]$ degrees to the observed angles of the two motors on the front left leg.

\textbf{4) Carrying an object.} All tasks above involves adapting to changes in dynamics only. In this task, we design a scenario where both dynamics and reward changes. Specifically, we ask the robot to carry a box of $1$ Kg while running forward. The new reward is how far the box is carried without falling to the ground. This task stresses the need of adapting the behavior of the policy, and a robust policy with a single behavior is unlikely to succeed.

\newtext{Note that the testing variations in the simulated tasks (slope, motor offset, object) are not included during the training of the policy. The policy needs to adapt to this novel task by leveraging the latent space acquired for the diverse set of dynamics seen during training. } For all simulated tasks except for extended randomization range, we also need to determine what values to use for the parameters randomized during training. As there is no single set of values that is representative of the robot, we also randomize these parameters using the same training range (Table \ref{tbl:randomization_range}) for those tasks.

\begin{table}[b]
 \vspace{-3mm}
%   \vspace{-2mm}
\caption{Randomized parameters and their range used in training.}
\vspace{-4mm}
\begin{center}
\begin{tabular}{|c|c|c|}
    \hline
    \bf{parameter} & \bf{lower bound}  & \bf{upper bound} \\ \hline \hline
    mass & 60\% & 160\%   \\  \hline
	motor friction & 0.0Nm & 0.2Nm   \\  \hline
	inertia & 25\% & 200\%   \\  \hline
	motor strength & 50\% & 150\%   \\  \hline
	latency & 0ms & 80ms \\  \hline
	battery voltage & 10V & 18V   \\  \hline
	contact friction & 0.2 & 1.25   \\  \hline
	joint friction & 0.0Nm & 0.2Nm   \\  \hline
\end{tabular}
\end{center}
\label{tbl:randomization_range}
\vspace{-4mm}
\end{table}

\subsection{Results on real robot}

\begin{figure}[t]
    \centering
    \includegraphics[width=0.42\textwidth]{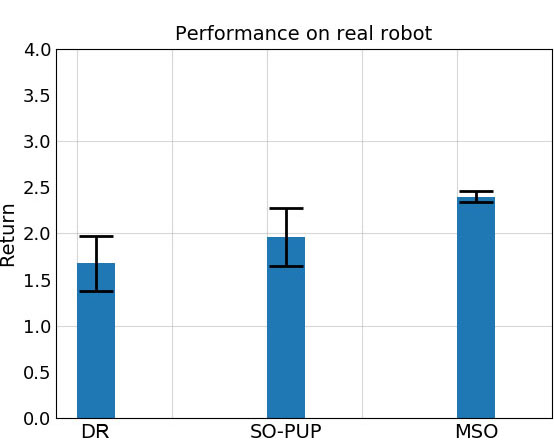}
    \caption{\small Sim-to-real performance comparison on the Minitaur robot \newtext{(corresponding to Task 1: Sim-to-real transfer as described in \ref{ssec:tasks})}. Error bar denotes on standard deviation.}
    \vspace{-2mm}
	\label{fig:perf_sim2real}
\end{figure}

We evaluate MSO on real Minitaur robot for the three tasks described in Section \ref{ssec:tasks}. For MSO and the baseline methods, we use the policy with the highest training performance among the five trials to deploy on the real hardware. For MSO and SO-PUP, we allow $15$ episodes for the adaptation and repeat the best policy for three times to obtain the final performance. For the sim-to-real task, we evaluate all three methods and report the result in Figure \ref{fig:perf_sim2real}. We see that MSO is able to not only achieve a better performance on average, but also obtain lower variance in performance.

\begin{figure}[b]
    \centering
    \includegraphics[width=0.42\textwidth]{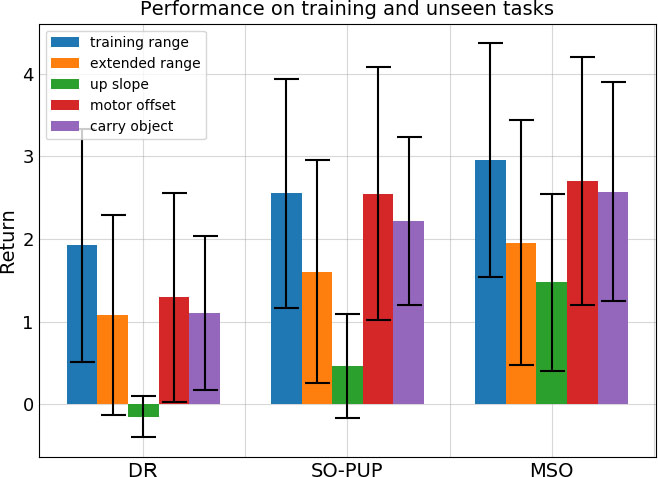}
    \caption{\small Comparison of performance on the training randomization range and generalization to unseen tasks. Error bar denotes on standard deviation. }
	\label{fig:perf_sim}
\end{figure}

\begin{figure}
    \centering
    \includegraphics[width=0.48\textwidth]{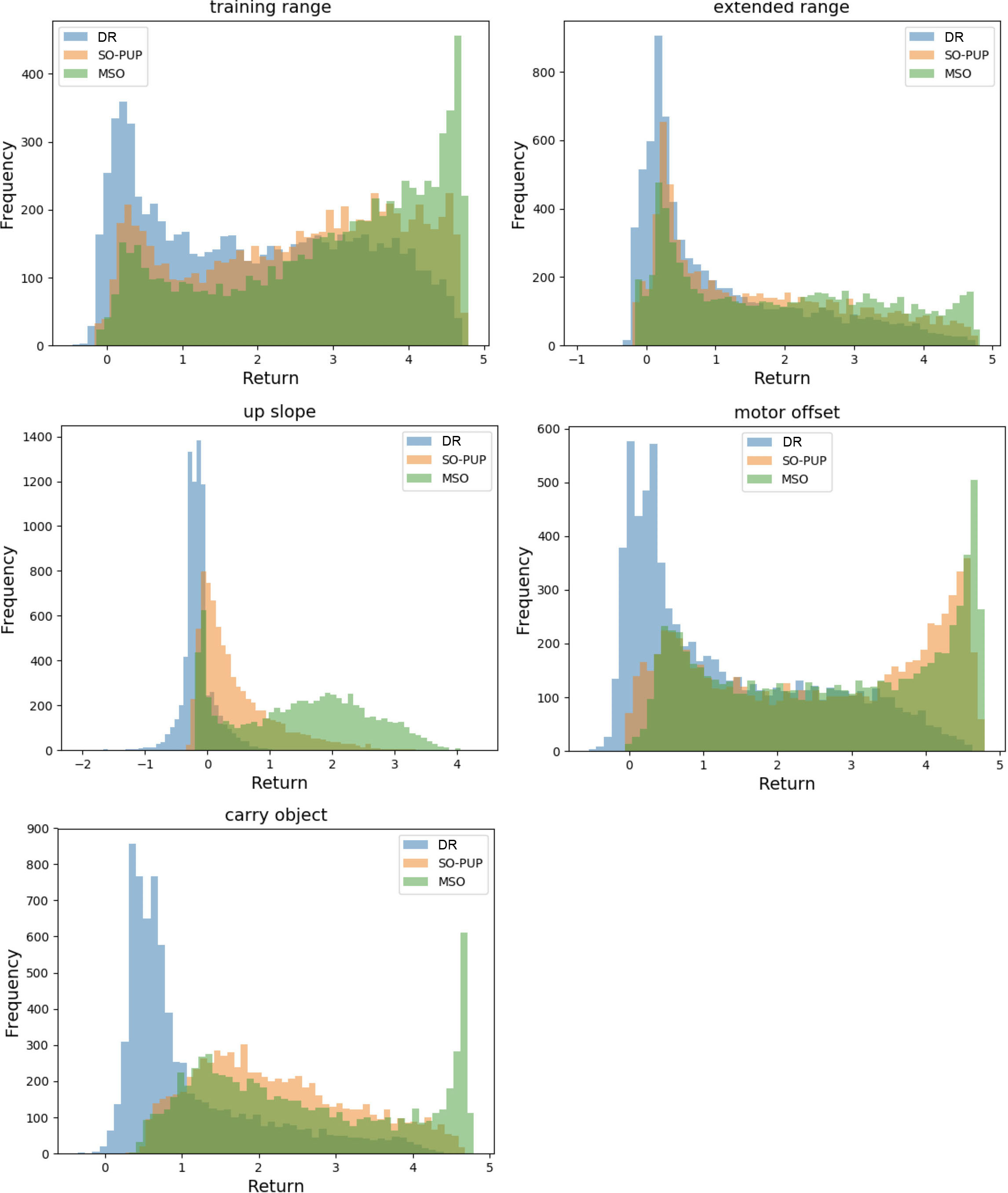}
    \vspace{-4mm}
    \caption{\small \newtext{Histograms for the returns of the sampled tasks in different adaptation problem. Each method was evaluated on $7,500$ sampled tasks for each adaptation problem.}}
	\label{fig:histogram}
	\vspace{-2mm}
\end{figure}

\begin{figure*}
    \centering
    \includegraphics[width=\textwidth]{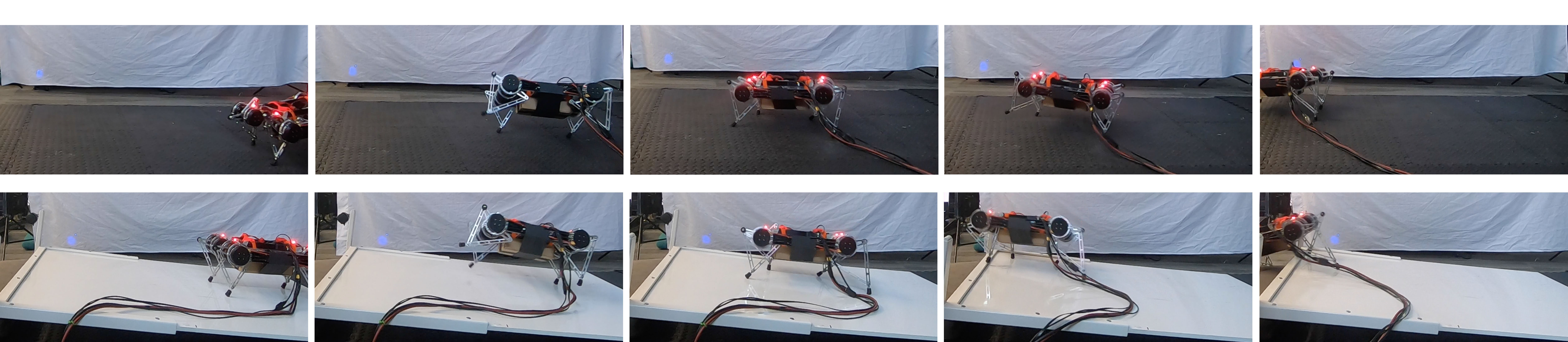}
    \caption{\small  Policy trained by MSO adapts to new tasks: front right leg weakened (top), walking up a slope (bottom).}
	\label{fig:real_minitaur}
	\vspace{-2mm}
\end{figure*}

For the task of weakened motor and slope climbing, we compare MSO to DR  \newtext{and SO-PUP}. As seen in the supplement video, when the front right leg is weakened, the robot lacks the strength to lift it up, and MSO finds a strategy that drags the front right leg forward without falling. On the other hand, DR still assumes full strength of the front right leg and relies on it to lift the base of the robot up, leading to it losing balance. Similarly for the task of climbing up the hill, MSO is able to find a strategy that successfully take the robot up the hill and go beyond the slope, while DR leads to the robot falling backward as it has only seen flat ground. \newtext{SO-PUP is able to learn different strategies that allow it to perform sim-to-real transfer to some extent, yet the resulting strategies are not rich enough to overcome the novel tasks.}

\subsection{More analysis in simulation}

We evaluate our method in simulated adaptation tasks to provide a more comprehensive analysis of our algorithm. \original{We first evaluate the training performance of MSO by testing it on the dynamics within the training range. As shown in Figure \ref{fig:perf_sim}, MSO notably outperforms the other two methods.} \newtext{We evaluate the performance of MSO and the baseline methods by testing them on the dynamics within the training range, as well as on the four adaptation tasks described in Section \ref{ssec:tasks}: extended randomization, climbing up a slope, biased motor zero position, and carrying an object.}

\original{We also report the performance of MSO and the baseline methods on the four adaptation tasks described in Section \ref{ssec:tasks}: extended randomization, climbing up slope, biased motor zero position, and carry object. All results can be seen in Figure \ref{fig:perf_sim}.}

\newtext{Figure \ref{fig:perf_sim} shows the mean and standard deviation for the three methods on different adaptation tasks. The statistics for each experiment are computed from $7,500$ samples. We also plot the histograms for the returns for each set of experiment to understand the reward distributions over a wide range of tasks (Figure \ref{fig:histogram}).} For all adaptation tasks, MSO is able to outperform both SO-PUP and DR. Notably, for the task of climbing up a slope, MSO achieved a clear advantage over the baseline methods, while DR is not able to achieve positive return. On the other hand, the difference between MSO and SO-PUP is smaller when an offset is added to the observed motor angle, while DR performs much worse. These results suggest that some tasks, such as climbing up the slope, are more sensitive to \original{learning a good latent strategy space}\newtext{latent space qualities} than other tasks. MSO also works well for the task of carrying the object, where the policy needs to adapt to changes in both dynamics and reward. As seen in the supplement video, MSO can successfully find a strategy that stabilizes the base of the robot to prevent the object from falling to the ground, while the baseline methods achieves worse performance.

\begin{figure*}
    \centering
    \includegraphics[width=0.8\textwidth]{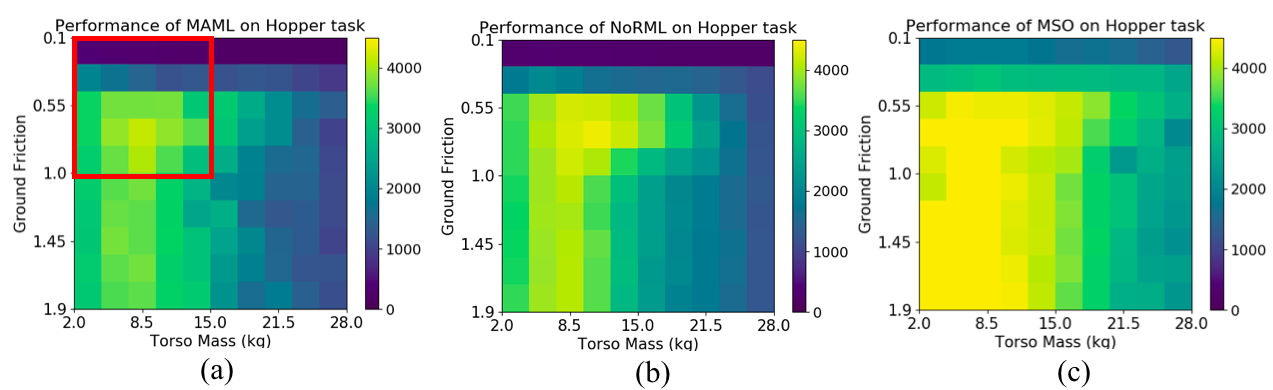}
    \vspace{-2mm}
    \caption{\small \newtext{Performance comparison between MAML (a), NoRML (b), and MSO (c) on the Hopper task. The squared region in (a) denotes the range of the training dynamics for all three methods. The color in the plot represents the performance of a task. The better the method performs with the dynamics parameters setting, the lighter the grid color is.}}
	\label{fig:metalearning_compare}
	\vspace{-2mm}
\end{figure*}

\subsection{Ablation study}

\begin{table}[b]
 \vspace{-3mm}
%   \vspace{-2mm}
\caption{Ablation study for the MSO algorithm.}
\vspace{-8mm}
\begin{center}
\begin{tabular}{|c|c|c|}
    \hline
    \bf{parameters} & \bf{mean return (training)} & \bf{mean return (extended)} \\ \hline \hline 
    $e$=$25$, $l$=$2$, $h$=$30$ & 2.95 &1.95 \\ \hline
    $e$=$15$ & 2.91 &1.85 \\ \hline 
    $e$=$1$ & 2.36 &1.51 \\ \hline \hline
    $e$=$25$, $l$=$2$, $h$=$30$ & 2.95 &1.95 \\ \hline
    $l$=$1$ & 2.84 &1.85 \\ \hline
    $l$=$5$ & 2.97 &1.95 \\ \hline \hline
    $e$=$25$, $l$=$2$, $h$=$30$ & 2.95 &1.95 \\ \hline
    $h$=$15$ & 2.70 &1.78 \\ \hline
    $h$=$50$ & 3.01 &1.94 \\ \hline
\end{tabular}
\end{center}
\label{tbl:ablation}
\vspace{-4mm}
\end{table}

We investigate how sensitive our algorithm is to different choices of hyper-parameters. In particular, we vary three key parameters for MSO: 1) $e$: the number of episodes allowed in SO during training, 2) $l$: the dimension of the latent space, and 3) $h$: the number of iterations between each SO during training. Our nominal model uses $e=25$, $l=2$ and $h=30$ for the three parameters. We vary one parameter at a time from the nominal setting and pick two values for each parameter being ablated. We test all variations of MSO on the training performance and the extended randomization task with $7,500$ samples each. During testing, we allow $15$ episodes for adaptation for all variations. Table \ref{tbl:ablation} shows the result of the ablation.

In general, our method is not very sensitive to different hyper-parameters. Interestingly, even when a single episode is allowed for SO during training, i.e. a random strategy is selected, the resulting policy can still outperform DR notably. This is possibly because training a policy in this setting is similar to training a set of DR policies with different random seeds, and during testing, the best performing one will be picked.
%\sehoon{Is SO(1) equal to DR? Then how does it outperform DR?} \wenhao{It's more similar to training multiple DRs and pick the best performing one during testing. Reworded.}

% \begin{figure}
%     \centering
%     \includegraphics[width=0.45\textwidth]{images/ablation_study}
%     \caption{\small Ablation study for the MSO algorithm. Nominal means the main setting used in the experiments. $n$ shots means the number of episodes allowed in SO during training. $dim(c)$ is the dimension of the latent space, and $h$ is the number of policy updates before SO is performed.}
% 	\label{fig:ablation}
% \end{figure}

% \begin{table}[b]
%  \vspace{-3mm}
% %   \vspace{-2mm}
% \caption{Ablation study for the MSO algorithm.}
% \vspace{-2mm}
% \begin{center}
% \begin{tabular}{|c|c|c|}
%     \hline
%     \bf{parameters} & \bf{mean return (training)} & \bf{mean return (extended)} \\ \hline \hline 
%     $e$=$25$, $l$=$2$, $h$=$30$ & 2.95 &1.95 \\ \hline \hline
%     $e$=$\mathbf{15}$, $l$=$2$, $h$=$30$ & 2.91 &1.85 \\ \hline 
%     $e$=$\mathbf{1}$, $l$=$2$, $h$=$30$ & 2.36 &1.51 \\ \hline \hline
%     $e$=$25$, $l$=$\mathbf{1}$, $h$=$30$ & 2.84 &1.85 \\ \hline
%     $e$=$25$, $l$=$\mathbf{5}$, $h$=$30$ & 2.97 &1.95 \\ \hline \hline
%     $e$=$25$, $l$=$2$, $h$=$\mathbf{15}$ & 2.70 &1.78 \\ \hline
%     $e$=$25$, $l$=$2$, $h$=$\mathbf{50}$ & 3.01 &1.94 \\ \hline
% \end{tabular}
% \end{center}
% \label{tbl:ablation}
% \vspace{-4mm}
% \end{table}

\subsection{\newtext{Comparison to gradient-based meta learning}}
\label{maml_compare}

\newtext{Finally, we evaluate our method on a completely different domain, training a simulated Hopper robot to hop forward, to compare it against two gradient-based meta learning algorithms: MAML \cite{finn2017model} and NoRML \cite{yang2019norml}. We evaluate all methods on dynamics that extend the training range by $100\%$, which are shown in Figure \ref{fig:metalearning_compare}. We observe that MSO significantly outperforms the other two methods both in terms of the training and generalization performance. We believe this is due to that MSO optimizes the latent input $\vc{c}$ in a low dimensional space (2D in our case), which allows more sample-efficient adaptation than gradient-based adaptation methods that need to adjust the entire policy network.}

%% file: conclusion.tex
\section{Discussion and Conclusion}

We have presented a learning algorithm for training locomotion policies that can quickly adapt to novel environments that are not seen during training time. The key idea to our method, Meta Strategy Optimization (MSO), is a meta-learning process that learns a latent strategy space suitable for fast adaptation during training, and quickly searches a good strategy to adapt to new rewards and dynamics during testing. We demonstrate MSO on a variety of simulated and real-world adaptation tasks, including walking on a slope, weakened motor, and carrying objects. MSO can successfully adapt to the novel tasks in $15$ episodes and outperforms other baseline methods.

Though MSO can successfully transfer policies to environments that are notably different from the training environments, it assumes that the testing environment does not change significantly over time. This restricts the type of tasks that MSO can be applied to. For example, if the robot needs to walk across an slippery surface and a rough surface, it would require changing the strategy when the surface type changes. One possible future direction to address this issue is to adopt the idea of hierarchical RL \cite{bacon2017option, liu2017learning} by treating the MSO-trained policy as a lower-level policy and train a higher-level policy that outputs the strategy. This will also enable the policy to adapt in an online fashion.